\documentclass[10pt,twocolumn,letterpaper]{article}

\usepackage{cvpr}              

\usepackage{graphicx}
\usepackage{amsmath}
\usepackage{amssymb}
\usepackage{booktabs}

\usepackage{amsfonts}
\usepackage{multirow}
\usepackage{pifont}
\usepackage{stfloats}
\usepackage{bm}
\usepackage{algorithm}  
\usepackage{algorithmicx}  
\usepackage{algpseudocode}
\usepackage{color}

\usepackage[pagebackref,breaklinks,colorlinks]{hyperref}

\usepackage[capitalize]{cleveref}
\crefname{section}{Sec.}{Secs.}
\Crefname{section}{Section}{Sections}
\Crefname{table}{Table}{Tables}
\crefname{table}{Tab.}{Tabs.}

\def\cvprPaperID{2403} 
\def\confName{CVPR}
\def\confYear{2023}

\usepackage[accsupp]{axessibility}
\begin{document}

\title{Trajectory-Aware Body Interaction Transformer for Multi-Person Pose Forecasting}

\author{Xiaogang Peng,
Siyuan Mao,
Zizhao Wu\footnotemark[2]\\
Department of Digital Media Technology, Hangzhou Dianzi University, Hangzhou, China \\
\tt\small $\{$pengxiaogang,siyuanmao,wuzizhao$\}$@hdu.edu.cn}



\twocolumn[{%
\renewcommand\twocolumn[1][]{#1}%
\vspace{-1.5cm}
\maketitle

\vspace{-1cm}
\begin{center}
\centering
\includegraphics[width=1.0\textwidth]{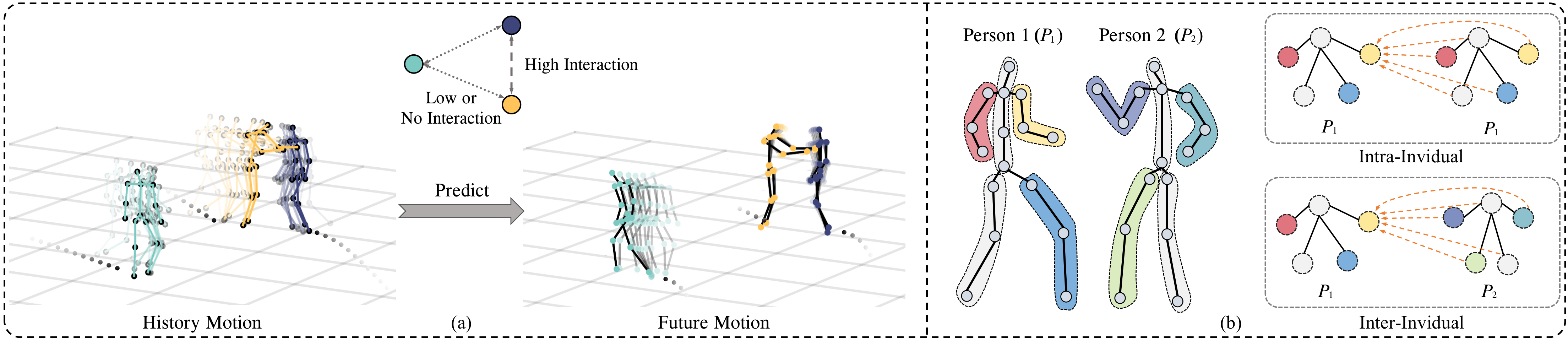} 
\captionof{figure}{(a) In complex crowd scenarios, different people may interact with one another at varying levels (low and high interactions) and at different positions (\ie, between near and far distances). (b) The illustration of our main idea on body part interactions. We divide the body joints into 5 parts, and the Intra-Individual branch is used to explore part relationships for each individual and the Inter-Individual branch aims to capture interaction dependencies of body parts between individuals. Our TBIFomer facilitates to  model body part interactions for intra- and inter-individuals simultaneously.}
\label{fig1}
\end{center}
}]

\renewcommand{\thefootnote}{\fnsymbol{footnote}}
\footnotetext[2]{Corresponding author.}

\begin{abstract}
\vspace{-3mm}
Multi-person pose forecasting remains a challenging problem, especially in modeling fine-grained human body interaction in complex crowd scenarios. Existing methods typically represent the whole pose sequence as a temporal series, yet overlook interactive influences among people based on skeletal body parts. In this paper, we propose a novel \textbf{T}rajectory-Aware \textbf{B}ody \textbf{I}nteraction Trans\textbf{former} (\textbf{TBIFormer}) for multi-person pose forecasting via effectively modeling body part interactions. Specifically, we construct a Temporal Body Partition Module that transforms all the pose sequences into a Multi-Person Body-Part sequence to retain spatial and temporal information based on body semantics. Then, we devise a Social Body Interaction Self-Attention (SBI-MSA) module, utilizing the transformed sequence to learn body part dynamics for inter- and intra-individual interactions. Furthermore, different from prior Euclidean distance-based spatial encodings, we present a novel and efficient Trajectory-Aware Relative Position Encoding for SBI-MSA to offer discriminative spatial information and additional interactive clues. On both short- and long-term horizons, we empirically evaluate our framework on CMU-Mocap, MuPoTS-3D as well as synthesized datasets (6 $\sim$ 10 persons), and demonstrate that our method greatly outperforms the state-of-the-art methods. Codes will be released  at \href{https://github.com/xiaogangpeng/TBIFormer}{https://github.com/xiaogangpeng/TBIFormer.}
\end{abstract}

\section{Introduction}
\label{sec:intro}

Recent years have seen a proliferation of work on the topic of human motion prediction \cite{c:7,c:15,c:17,c:19,c:20,c:22,c:23,c:24}, which aims to forecast future poses based on past observations. Similarly, understanding and forecasting human motion plays a critical role in the field of artificial intelligence and computer vision, especially for robot planning, autonomous driving, and video surveillance \cite{c:21,c:25,c:26,c:16}.
Although encouraging progress has been achieved, the current methods are mostly based on local pose dynamics forecasting without considering global position changes of body joints (global body trajectory) and often tackle the problem of single humans in isolation while overlooking human-human interaction. Actually, in real-world scenarios, each person may interact with one or more people, ranging from low to high levels of interactivity with instantaneous and deferred mutual influences \cite{b:1, c:32}. As illustrated in \cref{fig1} (a), two individuals are pushing and shoving with high interaction, whilst a third individual is strolling with no or low interaction. Thus, accurately forecasting pose dynamics and trajectory and comprehensively considering complex social interactive factors are imperative for understanding human behavior in multi-person motion prediction. However, existing solutions do not efficiently address these challenging factors. For example, Guo \etal \cite{c:14} propose a collaborative prediction task and perform future motion prediction for only two interacted dancers, which inevitably ignores low interaction influence on one’s future behavior. Wang \etal \cite{c:3} use local and global Transformers to learn individual motion and social interactions separately in a crowd scene. The aforementioned methods ignore the interactive influences of body parts and only learn temporal and social relationships without modeling fine-grained body interaction, which makes it difficult to capture complex interaction dependencies.

To solve this issue, we propose a novel Transformer-based framework, termed TBIFormer, which consists of multiple stacked TBIFormer blocks and a Transformer decoder. In particular, each TBIFormer block contains a Social Body Interaction Multi-Head Self-Attention (SBI-MSA) module, which aims at learning body part dynamics across inter- and intra-individuals and capturing fine-grained skeletal body interaction dependencies in complex crowd scenarios as shown in \cref{fig1} (b). More specifically, SBI-MSA learns body parts dynamics across temporal and social dimensions by measuring motion similarity of body parts rather than pose similarity of the entire body. In addition, a Trajectory-Aware Relative Position Encoding is introduced for SBI-MSA as a contextual bias to provide additional interactive clues and discriminative spatial information, which is more robust and accurate than the Euclidean distance-based spatial encodings.

In order to feed the TBIFormer a pose sequence containing both temporal and spatial information, an intuitive way is to retain body joints in time series. However, this strategy will suffer from noisy joints caused by noisy sensor inputs or inaccurate estimations. In this work, we propose a Temporal Body Partition Module (TBPM) that, based on human body semantics, transforms the original pose sequence into a new one, enhancing the network's capacity for modeling interactive body parts. Then, we concatenate the transformed sequences for all people one by one to generate a Multi-Person Body Part (MPBP) sequence for input of TBIFormer blocks, which enables the model to capture dependencies of interacting body parts between individuals. TBIFormer makes MPBP sequence suitable for motion prediction by utilizing positional and learnable encodings to indicate to whom each body part and timestamp belongs.

Finally, a Transformer decoder is used to further consider the relations between the current and historical context across individuals' body parts toward predicting smooth and accurate multi-person poses and trajectories. For multi-person motion prediction (with 2 $\sim$ 3 persons), we evaluate our method on multiple datasets, including CMU-Mocap \cite{c:1} with UMPM \cite{c:5} augmented and MuPoTS-3D \cite{c:4}. Besides, we extend our experiment by mixing the above datasets with the 3DPW \cite{c:2} dataset to perform prediction in a more complex scene (with 6 $\sim$ 10 persons). Our method outperforms the state-of-the-art approaches for both short- and long-term predictions by a large margin, with 14.4$\%$ $\sim$ 16.5$\%$ accuracy improvement for the short-term ($\le$ 1.0s) and 6.5$\%$ $\sim$ 18.2$\%$ accuracy improvement for the long-term (1.0s $\sim$ 3.0s). 

To summarize, our key contributions are as follows: 1) We propose a novel Transformer-based framework for effective multi-person pose forecasting and devise a Temporal Body Partition Module that transforms the original pose sequence into a Multi-Person Body-Part sequence to retain both temporal and spatial information. 2) We present a novel Social Body Interaction Multi-Head Self-Attention (SBI-MSA) that learns body part dynamics across inter- and intra-individuals and captures complex interaction dependencies. 3) A novel Trajectory-Aware Relative Position Encoding is introduced for SBI-MSA to provide discriminative spatial information and additional interactive clues. 4) On multiple multi-person motion datasets, the proposed TBIFormer significantly outperforms the state-of-the-art methods.

\section{Related Work}
\subsection{Single-Person Pose Forecasting}Predicting human motion offers enormous promise for surveillance, autonomous driving, and human-robot interaction. Although recurrent neural networks (RNNs) have shown advantages in processing this typical sequence-to-sequence problem \cite{c:15,c:36,c:37}, discontinuity and error accumulation often happen due to the frame-by-frame prediction manner. To address these issues, some feed-forward networks such as graph convolution networks (GCNs) and temporal convolution networks (TCNs) are used to explore spatial and temporal dependencies \cite{c:19,c:20,c:29,c:31,c:51}. Besides, Mao \etal \cite{c:12} introduce an attention-based feed-forward network to capture the similarity between the current motion context and the historical motion sub-sequences and process the result via GCNs for long-term prediction. All the above methods only model local pose dynamics, ignoring global body translation and inter-individual body interaction. However, learning both local and global pose dynamics and modeling fine-grained human-human interaction are essential for comprehending human behavior in a complex 3D environment \cite{c:3,c:32}.

\begin{figure*}[t]
\centering
\includegraphics[width=0.9\textwidth]{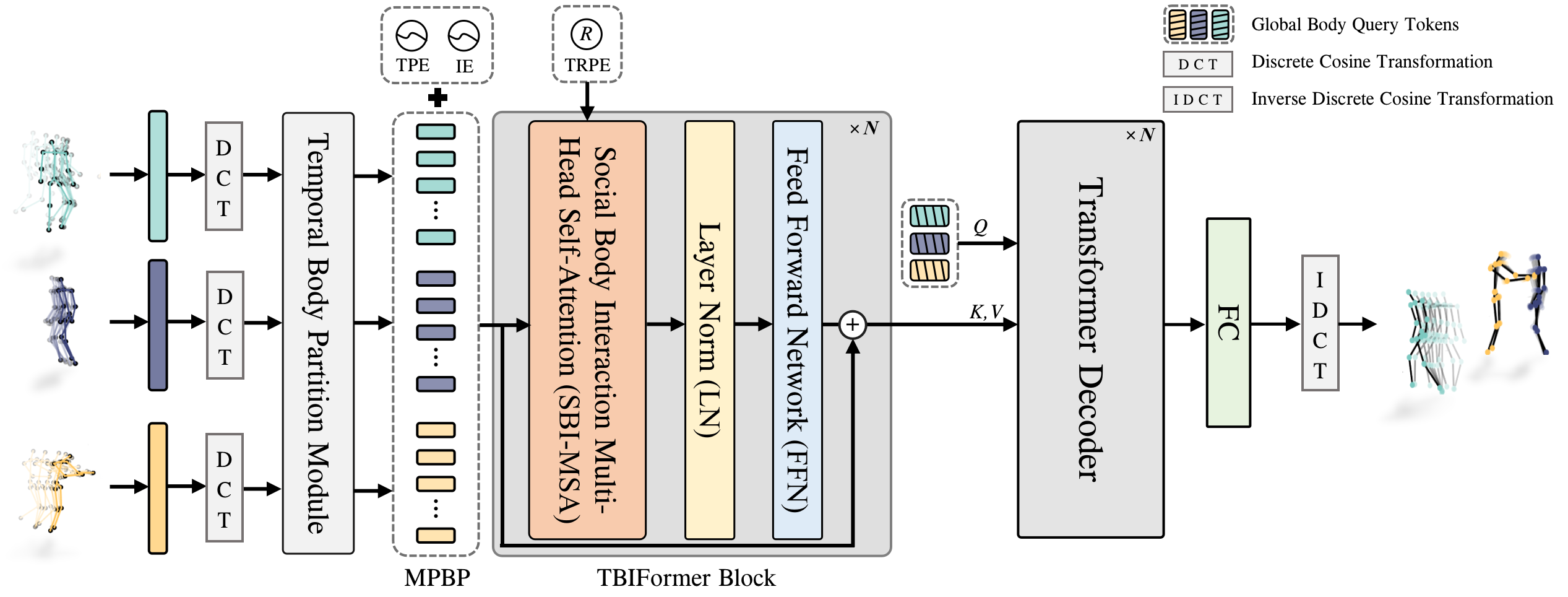} 
\caption{Overview of the proposed TBIFormer framework. Given the observed pose sequences of 3 persons, TBIFormer transforms them into displacement sequences as input and then forecasts future poses for each person. At the head and tail of TBIFormer, we adopt a Discrete Cosine Transformation (DCT) \cite{c:10} that discards the high-frequency information for a more compact representation in displacement trajectory space \cite{c:9}.}
\label{fig2}
\end{figure*}

\subsection{Multi-Person Pose Forecasting} In order to address fine-grained human-human interaction, some recent approaches are proposed for multi-person pose and trajectory forecasting. For example, Adeli \etal \cite{c:13} propose to combine scene context and use graph attention networks to model interaction between humans and objects. Guo \etal \cite{c:14} present a collaborative prediction task and use a two-branch attention network for the prediction of two interacted persons. Wang \etal \cite{c:3} present a Transformer-based framework to forecast multi-person motion in a scenario with more people. Furthermore, this method produces unrealistic poses since they solely concentrate on individual and social modeling in the time dimension. Despite the novelty of these works, skeletal body interaction between individuals is not captured effectively. In this work, we propose TBIFormer that learns skeletal body part dynamics for intra- and inter-individuals to effectively capture complex interaction dependencies.

\subsection{Multi-Person Social Interaction} Pedestrian trajectory prediction is a representative issue for multi-person social interaction. Existing methods for the task can be categorized based on how they model time and social dimensions. RNNs \cite{c:40} and Transformers \cite{c:41} are the preferred models \cite{c:42,c:43,c:44} to process the trajectory sequence for temporal modeling, and graph neural networks (GNNs) \cite{c:45} are often adopted as social models for interaction modeling \cite{c:46,c:47,c:52,c:53}. While performing well, these studies only focus on individuals' global movement without modeling detailed human joint dynamics.  In this work, we investigate our TBIFormer to consider fine-grained human-human interaction via modeling skeletal body part dynamics among individuals and predict future motion for 3 $\sim$ 10 persons in 3D scenes.


\section{Method}
In this section, we introduce our Trajectory-Aware Body Interaction Transformer (TBIFormer), which contains multiple stacked TBIFormer blocks and a Transformer decoder followed by fully connected layers, as shown in \cref{fig2}. Each TBIFormer block has a Social Body Interaction Multi-Head Self-Attention (SBI-MSA) module for modeling body part interactions across temporal and social dimensions. The proposed TBIFormer is also equipped with a Temporal Body Partition Module (TBPM), which aims to better learn body parts' spatial and temporal information within the skeletal sequences. In addition, temporal positional encoding, person identity encoding, and trajectory-aware relative position encoding are introduced to preserve time, identity, and discriminative spatial information. In the following, the problem definition and our key modules are described in detail.

\subsection{Problem Definition}
Supposing the observed skeletal poses from person $p$ are $X_{1:T+1}^p=\{x_1^p,x_2^p,...,$ $x_{T+1}^p\}$ with $T+1$ frames, where $p = 1,2, ... P$. For simplicity, we omit subscript $p$ when $p$ only represents an arbitrary person, \eg, taking $x^p_{1:t}$ as $x_{1:t}$. Instead of absolute joint positions in the world coordinate, we use ${y}_i = x_{i+1} - x_{i}$ to obtain instantaneous pose displacement at time $i$, which will provides more valuable dynamics information \cite{c:7, c:3}. The whole displacement sequence is defined as $Y_{1:T} = \{{y_1,y_2,...,y_{T}}\}$. Given the displacement sequence $Y_{1:T}$ of each person, our goal is to predict the $N$ frames of future displacement trajectory $Y_{T+1:T+N}$ and transform it back to the pose space $X_{T+2:T+N+1}$. 

\begin{figure}[t]
\centering
\includegraphics[width=0.5\textwidth]{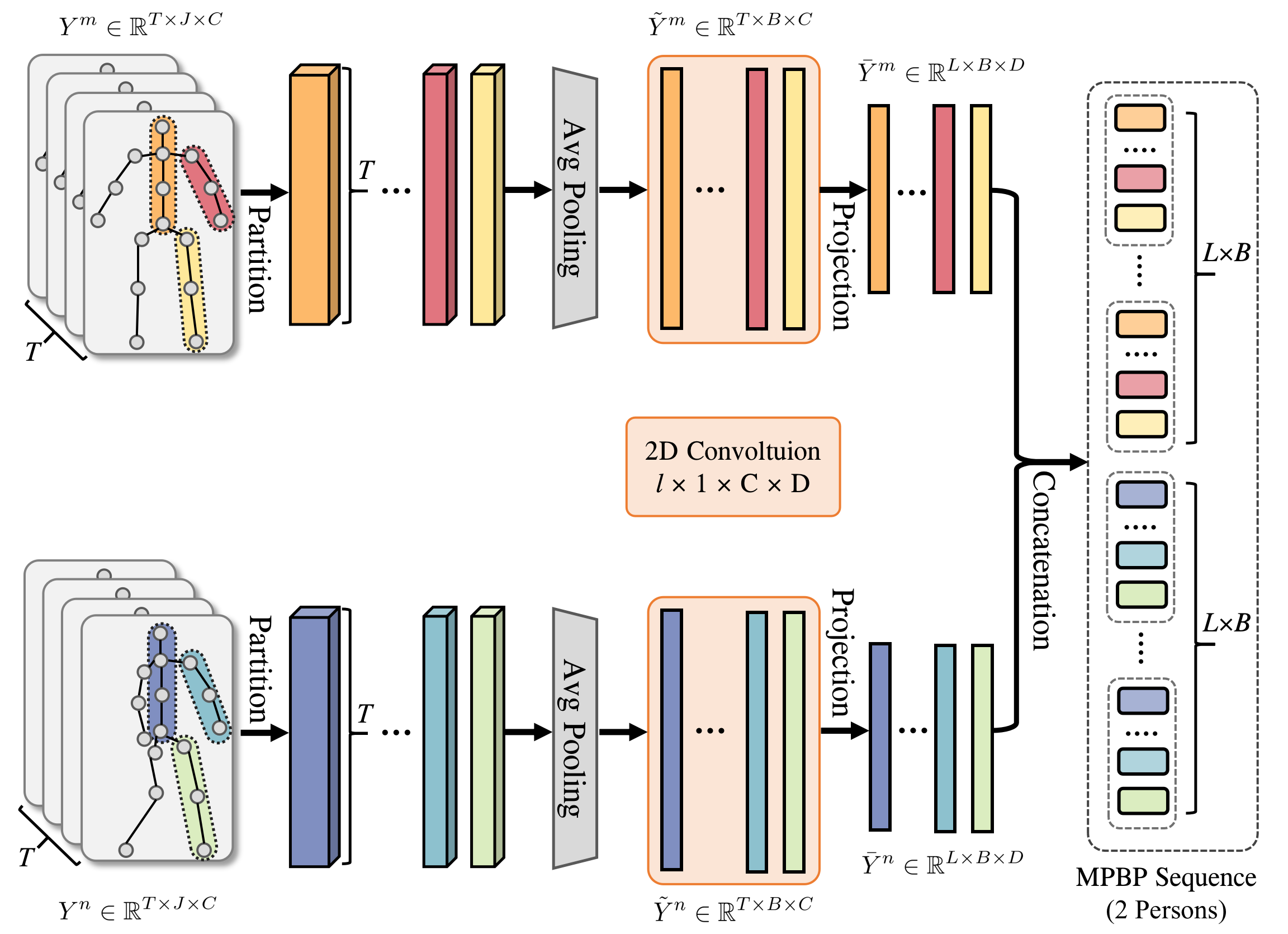} 
\caption{The illustration of the proposed Temporal Body Partition Module (TPBM). TPBM performs three main operations (\ie, partition, projection, and concatenation) on the input pose sequence to generate a Multi-Person Body-Part (MPBP) sequence.}
\label{fig3}
\end{figure}

\subsection{Temporal Body Partition Module}
To better retain both spatial and temporal information of the skeleton sequences, we propose TBPM that first transforms the pose sequence of each person into a sequence that contains body parts during a short time period. Then, TBPM concatenates all the transformed individuals' sequences into a Multi-Person Body-Part (MPBP) sequence for the following Transformers. There are three primary processes in TBPM, \ie, partition, projection and concatenation, which are described below. 

\noindent\textbf{Partition.} Given the displacement sequence of the $p$-th person $Y^{p} \in{\mathbb{R}^{T\times J\times C}}$, where $T$ and $J$ represent the numbers of frames and joints, respectively, and $C=3$ represents the dimension of the 3D coordinates, we first divide the sequence into $B=5$ body parts (\eg left and right arms, left and right legs, core torso) based on natural human skeletal structure, and then down-sample each body part by average-pooling. After the above operations, the sequence is represented as $\tilde{Y}^{p}\in{\mathbb{R}^{T\times B\times C}}$.

\noindent\textbf{Projection.} The goal of the projection operation is to initially extract spatial and temporal information. Specifically, we use 2D convolution with a kernel size of $l\times 1$ on $\tilde{Y}^{p}$ to obtain 2D feature map $\overline{Y}^{p} \in{\mathbb{R}^{L\times B\times D}}$,  where $L = \lfloor (T-l+1)/stride \rfloor$ and $D $ denote the number of output channels. We denote “$padding$” and “$stride$” as the padding size and stride size of the convolutional filter. 

\noindent\textbf{Concatenation.} Following projection, the encoding of all the $B$ body parts are concatenated for all the $L$ timesteps to form a new sequence with the length of $U = L\times B$. Next, we concatenate the sequences of all the $P$ persons one by one for a merged Multi-Person Body-Part (${\rm MPBP} \in \mathbb{R}^{M\times D}$) sequence, where $M$ denotes $P\times U$. MPBP sequence allows our TBIFormer to learn individuals' body part dynamics across temporal and social dimensions.

\subsection{Temporal Positional and Person Identity Encoding}
Similar to the original Transformer \cite{c:41}, we apply sinusoidal positional encoding to convey to TBIFormer the timestep associated with each element in the MPBP sequence. Instead of encoding the position of each element based on index in the whole MPBP sequence, we first compute timestamp features based on the timesteps of each person and obtain temporal positional encoding $\tau_{p}\in{\mathbb{R}^{T\times d_{\tau}}}$, where $d_{\tau}$ is the feature dimension of the timestamp. Then we utilize the interleaved repeating function to repeat the encoding elements for B body parts and concatenate the encoding of all individuals. The final temporal positional encoding (TPE) is formulated as $\hat{\tau}\in{\mathbb{R}^{M\times d_{\tau}}}$.

To provide identity information of each individual in the MPBP sequence, we also inject a learnable person identity encoding $	\nu\in{\mathbb{R}^{M\times d_{\nu}}}$, indicating which individual each element belongs to, where $d_{\nu}$ denotes the feature dimension. Notably, the identity encoding (IE) is randomly initialized and repeated for the time and body parts using the same repeating method for TPE.

\begin{figure}[h]
\centering
\includegraphics[width=0.4\textwidth]{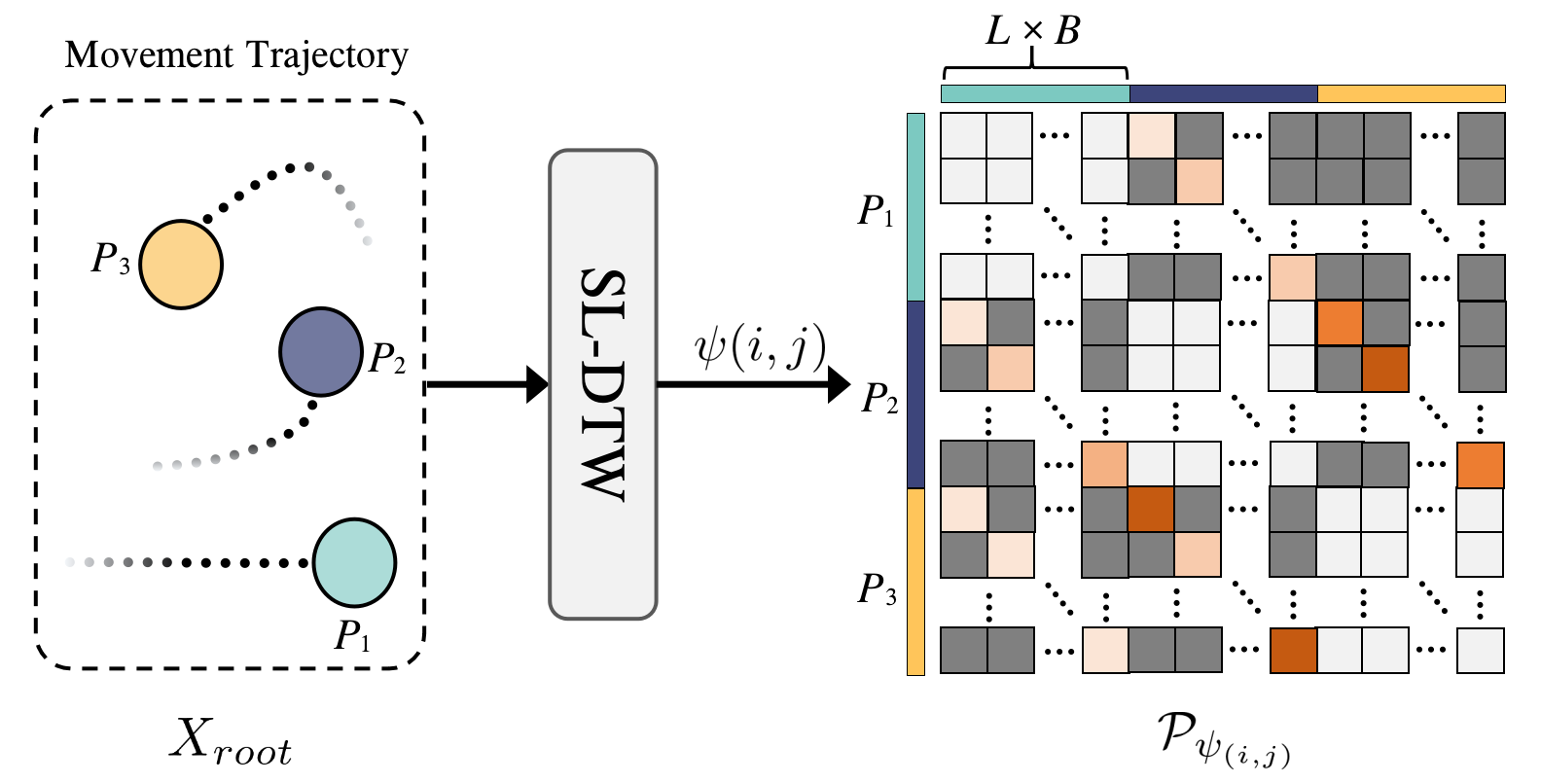} 
\caption{The overview of the proposed Trajectory-Aware Relative Position Encoding (TRPE).}
\label{fig4}
\end{figure}

\subsection{Trajectory-Aware Relative Position Encoding} An instinctive assumption is that the closer a few people are, the higher the interaction they may have. Yet, in the complex crowd situation, one person may have their back turned to a nearby individual with no interaction, or, as depicted in \cref{fig1}, a person may just pass by two interacting individuals, exhibiting low interaction yet close proximity. Therefore, Euclidean distance-based spatial position encodings struggle to provide discriminative spatial information and distinguish individuals who are actually interacting. 

In this paper, our observation is that people interacting in 3D space tend to move in the same or face-to-face direction, as opposed to the deviated direction. The main challenge is that directly calculating body orientation of the human skeleton data and the angle between the individuals is tedious and costly. To solve it, we find that movement trajectories can also provide vital information and circumvent the aforementioned limitations. Therefore, we propose a novel Trajectory-Aware Relative Position Encoding (TRPE) by measuring the similarity of movement trajectories, which can aggregate both corresponding movement pattern and spatial information. Dynamic Time Warping (DTW)\cite{c:49,c:50} is a more robust method to measure trajectory (series) similarity than Euclidean distance. In this work, we employ an efficient and differentiable algorithm variant called Soft-DTW  \cite{c:48}, which can be defined as,
\begin{equation}
\label{eq1}
\small{
    \begin{split}
    D(i,j)& = min\{ D(i, j-1), D(i-1,j), D(i-1,j-1)\}\\ 
           & + \delta(i,j),
    \end{split}
}
\end{equation}
where $D{(i,j)}$ denotes the shortest distance between sub-sequence $S1 = (s_{1}, s_{2}, ... , s_{i})$ and $S2 = (s_{1}, s_{2}, ... , s_{j})$ and $\delta{(\cdot,\cdot)}$ is differentiable cost function. In order to dynamically obtain trajectory similarity according to a certain timestep as opposed to complete timestamps, we propose a Shifted Local DTW (SL-DTW) mechanism based on Soft-DTW. Similar to the convolution operation, SL-DTW calculates the similarity between individuals at a specific window size and shifts step-by-step, which will provide more precise relative information. See \cref{alg1} for a detailed description of the SL-DTW process.

\begin{algorithm}[ht!]  
	\renewcommand{\algorithmicrequire}{\textbf{Input:}}
	\renewcommand{\algorithmicensure}{\textbf{Output:}}
	\caption{Shifted Local DTW mechanism (SL-DTW)}  
	\label{alg1}
	\begin{algorithmic}[1] 
		\Require   The root trajectory sequence of person $m$ and person $n$, $X_{r}^{m}=(x_{r,1}^{m}, x_{r,2}^{m}, ... , x_{r,T}^{m})$ and $X_{r}^{n}= (x_{r,1}^{n}, x_{r,2}^{n}, ... , x_{r,T}^{n})$; The size of local window and shift stride, $l$ and $stride$; The length of input sequence, $T$;
		\Ensure  The trajectory similarity $D^{<m,n>}$ between person $m$ and $n$;
            \State  $D^{<m,n>}= \left[\ \right]$
		\For{$i=0$; $i<\lfloor(T-l+1)/stride\rfloor$; $i+=stride$} 
		\State  $D^{<m,n>} = stack(D^{<m,n>}, D{(x_{(r,i+l)}^{m},x_{(r,i+l)}^{n})})$
		\EndFor 		
	\end{algorithmic}
\end{algorithm}

Given the trajectory similarity distance $\tilde{D} \in\mathbb{R}^{P\times L}$ among $P$ persons, we need to map the distance to an integer set for relative position encoding. The common way to address this issue is the clip function: $h(\tilde{D})=max(-\beta,min(\beta, \tilde{D}))$, which inevitably eliminates the context of long-distance relative position. Hence, we alternatively use the piecewise function \cite{c:11} $g(\cdot)$ that maintains long-range information for indexing relative distances to corresponding encodings, and then define the indexed matrix through the SL-DTW distance as follows:
\begin{equation}\label{eq2}
\small{
\psi{(i,j)}=\left\{
\begin{aligned}
g(\eta),\quad\quad\quad\quad\quad\quad\quad\quad\enspace\ \iota\neq \kappa,m\neq n,\\
g(0), \quad\quad\quad\quad\quad\quad\quad\quad\quad\quad\quad\enspace m=n,\\
g(\tilde{D}^{<m,n>}_{(\iota,\kappa)}),\quad\quad\quad\quad\quad\enspace  \iota=\kappa, m\neq n,\\
\end{aligned}
\right.
}
\end{equation}
where $\iota$ and $\kappa$ denote different timesteps from different person. In \cref{eq2}, to reduce additional computation, we ignore relations between person $m$ and $n$ on the condition of $\iota\neq \kappa$ and input a larger value $\eta$ in $g(\cdot)$ instead. The piecewise index function is presented as
\begin{equation}
\label{eq3}
\small{
g(e)=\left\{
\begin{aligned}
[e],\quad\quad\quad\quad\quad\quad\quad\quad\quad\quad\quad\quad |e|\le\alpha,\\
sign(e)\times \quad\quad\quad\quad\quad\quad\quad\quad\quad       |e|>\alpha, \\
min(\beta, [\alpha+\frac{\ln{(|e|/\alpha)}}{\ln{(\gamma/\alpha)}} (\beta-\alpha)]),\quad\quad\enspace
\end{aligned}
\right.
}
\end{equation}
where $[\cdot]$ is a round operation, $sign()$ determines the sign of a number, \ie, returning 1 for positive input, -1 for negative, and 0 for otherwise. $\alpha$ controls the piecewise point, $\beta$  limits the output in the range of $[- \beta, \beta]$, and $\gamma$ tunes the curvature of the logarithmic part. 

Finally, as shown in \cref{fig4}, we embed the indexed matrix $\psi{(i,j)}$ of trajectory similarity as our TRPE $ \mathcal{P}_{\psi(i,j)}\in \mathbb{R}^{M\times d_z}$ and denote $M$ = $P\times L\times B$, which are shared throughout all attention layers of SBI-MSA.

\subsection{SBI-MSA Module}In each TBIFormer block, we aim to construct a Social Body Interaction Multi-Head Self-Attention (SBI-MSA) module to effectively model body part dynamics for inter- and intra-individual. Given the motion features extracted by TBPM, SBI-MSA, based on motion-wise attention computation, can further optimize pose dynamics and capture complex body interaction dependencies among individuals. Let $H=[h_1^{},...,h_n]\in \mathbb{R}^{n\times d}$ denotes the input representation for attention module, where $d$ is the hidden dimension. SBI-MSA takes as input keys $K$, queries $Q$ and values $V$, each of which is projected by the corresponding parameter matrix $W_Q\in \mathbb{R}^{d\times d_z}$, $W_K\in \mathbb{R}^{d\times d_z}$ and $W_V\in R^{d\times d_z}$. The output of SBI-MSA is computed as
\begin{equation}\label{eq4}
\small{
Q=HW_Q,\quad K=HW_K,\quad V=HW_V,
}
\end{equation}
\vspace{-3mm}
\begin{equation}\label{eq5}
\small{
{\text{SBI-MSA}}(Q,K,V) = softmax(A)V.
}
\end{equation}

We integrate the TRPE $ \mathcal{P}_{\psi(i,j)}$ on the attention map to consider the interaction between individual dynamics features and spatial clues across temporal and social dimensions. Denoting $A_{ij}$ as the $(i,j)$-element of the Query-Key product matrix $ A $, we have 
\begin{equation}\label{eq6}
\small{
A_{ij} = \frac{Q_{i} \cdot K_j+ b_{i,j}^{\text{TRPE}}}{\sqrt{d_z}}, 
}
\end{equation}
\vspace{-3mm}
\begin{equation}\label{eq7}
\small{
b_{i,j}^{\text{TRPE}} = Q_{i} \cdot  \mathcal{P}_{\psi_{(i,j)}},
}
\end{equation}
where $b_{i,j}^{\text{TRPE}}$ is a contextual bias for the attention map.



\begin{table*}[t]
\centering
\renewcommand\arraystretch{0.9}
\resizebox{.9\linewidth}{!}{
\begin{tabular}{@{}clccccccccccccccccc@{}}
\toprule
\multicolumn{1}{l}{} & \multicolumn{1}{c}{} & \multicolumn{4}{|c}{\begin{tabular}[c]{@{}c@{}}CMU-Mocap (UMPM)\\ (3 persons)\end{tabular}} & \multicolumn{4}{|c}{\begin{tabular}[c]{@{}c@{}}MuPoTS-3D\\ (2 $\sim$3 persons)\end{tabular}} & \multicolumn{4}{|c}{\begin{tabular}[c]{@{}c@{}}Mix1\\ (6 persons)\end{tabular}} & \multicolumn{4}{|c}{\begin{tabular}[c]{@{}c@{}}Mix2\\ (10 persons)\end{tabular}} \\ \midrule
                     & Method & \multicolumn{1}{|c}{0.2s} & 0.6s & 1.0s & Overall & \multicolumn{1}{|c}{0.2s} & 0.6s & 1.0s & Overall & \multicolumn{1}{|c}{0.2s} & 0.6s & 1.0s & Overall & \multicolumn{1}{|c}{0.2s} & 0.6s & 1.0s & Overall \\
                     \midrule
\multirow{4}{*}{\rotatebox{90}{\textbf{JPE}}} 
& HRI\cite{c:12} & \multicolumn{1}{|c}{49} & 130 & 207 & 129 & \multicolumn{1}{|c}{81} & 211 & 323 & 205 & \multicolumn{1}{|c}{51} & 141 & 233 & 142 & \multicolumn{1}{|c}{52} & 140 & 224 & 139  \\
& MSR\cite{c:29} & \multicolumn{1}{|c}{53} & 146 & 231 & 143 &  \multicolumn{1}{|c}{79} & {222} & {374} & 225 & \multicolumn{1}{|c}{49} & 132 & 220 & 134 & \multicolumn{1}{|c}{60} & 153  & 243 & 152 \\
& MRT*\cite{c:3} & \multicolumn{1}{|c}{36} & {115} & {192} & 114 & \multicolumn{1}{|c}{78} & 225 & 349 & 217 & \multicolumn{1}{|c}{37} & {122} & {212} & 124 & \multicolumn{1}{|c}{38} & {126} & {214} & 126 \\
& Ours*  & \multicolumn{1}{|c}{\textbf{30}} & \textbf{109} & \textbf{182} & \textbf{107} & \multicolumn{1}{|c}{\textbf{66}} & \textbf{200} & \textbf{319} & \textbf{195} & \multicolumn{1}{|c}{\textbf{34}}  & \textbf{121} & \textbf{209} & \textbf{121} & \multicolumn{1}{|c}{\textbf{34}}  & \textbf{118}  & \textbf{198} & \textbf{117}  \\ \midrule

\multirow{4}{*}{\rotatebox{90}{\textbf{APE}}} 
& HRI\cite{c:12}  & \multicolumn{1}{|c}{41} & 97 & 130 & 89 &  \multicolumn{1}{|c}{70} & 136 & 174 & 127 & \multicolumn{1}{|c}{38} & 92 & 122 & 84 &  \multicolumn{1}{|c}{41} & 100 & 133 & 91 \\
& MSR\cite{c:29} & \multicolumn{1}{|c}{46} & 106 & 137 & 96 & \multicolumn{1}{|c}{71} & {148} & 190 & 136 & \multicolumn{1}{|c}{41} & 92 & 120 & 84 & \multicolumn{1}{|c}{48} & 110 & 148 & 102 \\
& MRT*\cite{c:3} & \multicolumn{1}{|c}{36}& {108} & {159} & 101 & \multicolumn{1}{|c}{71} & 166 & 217 & 151 &  \multicolumn{1}{|c}{36} & {109} & {166} & 104 & \multicolumn{1}{|c}{38} & {115} & {178} & 110  \\
& Ours*  & \multicolumn{1}{|c}{\textbf{27}} &  \textbf{84} &  \textbf{118} & \textbf{76} & \multicolumn{1}{|c}{\textbf{60}} & \textbf{132} & {\textbf{170}} & \textbf{121} & \multicolumn{1}{|c}{\textbf{28}} & \textbf{81}  & \textbf{113} & \textbf{74} & \multicolumn{1}{|c}{\textbf{30}} & \textbf{89}  & \textbf{124} & \textbf{81} \\ \midrule

\multirow{4}{*}{\rotatebox{90}{\textbf{FDE}}} 
& HRI\cite{c:12}  & \multicolumn{1}{|c}{31} & 90 & 158 & 93 &  \multicolumn{1}{|c}{63} & 173 & 279 & 172 & \multicolumn{1}{|c}{37} & 107 & 192 & 112  &  \multicolumn{1}{|c}{35} & 101 & 177 & 104 \\
& MSR\cite{c:29} & \multicolumn{1}{|c}{29} & 94 & 175 & 99 & \multicolumn{1}{|c}{58} & {184} & 335 & 192 & \multicolumn{1}{|c}{29} & 91 & 169 & 96 & \multicolumn{1}{|c}{38} & 113 & 185 & 112 \\
& MRT*\cite{c:3} & \multicolumn{1}{|c}{27}& {88} & {157} & 91 & \multicolumn{1}{|c}{59} & 187 & 309 & 185 &  \multicolumn{1}{|c}{29} & {100} & {189} & 106 & \multicolumn{1}{|c}{29} & {98} & {185} & 104  \\
& Ours*  & \multicolumn{1}{|c}{\textbf{18}} &  \textbf{72} &  \textbf{133} & \textbf{74} & \multicolumn{1}{|c}{\textbf{49}} & \textbf{163} & {\textbf{277}} & \textbf{163} & \multicolumn{1}{|c}{\textbf{23}} & \textbf{89}  & \textbf{168} & \textbf{93} & \multicolumn{1}{|c}{\textbf{21}} & \textbf{81}  & \textbf{151} & \textbf{84}  \\

\bottomrule
\end{tabular}
}
\caption{Results of JPE, APE and FDE (in mm) on different datasets. We compare our method with the previous SOTA methods for short-term and long-term predictions. Best results are shown in boldface. (* means multi-person motion prediction method.)}
\label{table1}
\end{table*}

\subsection{Transformer Decoder}
As illustrated in \cref{fig2}, we concatenate joint coordinates of the last observed sub-sequence (length = $l$) from each person for all the body joints and down-sample them on time dimension by 1D Convolution (kernel size = $l$) as global body query tokens. Key and value tokens are the output of the TBIFormer block. We utilize a standard Transformer decoder\cite{c:41} to encode the relations between the current (queries) and historical context (keys) across individuals. At the end of the decoder, we adopt two fully connected (FC) layers followed by an Inverse Discrete Cosine Transformation (IDCT) \cite{c:10} to generate the future motion trajectory $X_{{T+2}:T+N+1}$ for each individual.

\subsection{Loss Function}
We use a reconstruction loss based on the Mean Per Joint Position Error (MPJPE) for optimization. In particular, for one training sample, the loss is represented as
\vspace{-2mm}
\begin{equation}\label{eq8}
\small{
L_{rec} = \frac{1}{J*N}\sum_{i=N+1}^{T+N}{\sum_{j=1}^{J}{||\hat{y}_{i,j} - y_{i,j}||^2}},
}
\end{equation}
where $\hat{y}_{j,t}$ and $y_{j,t}$ are ground-truth and estimated pose displacement at time $i$. $J$ represents the number of body joints.


\begin{table}[t]
\renewcommand\arraystretch{0.9}
\centering
\resizebox{\linewidth}{!}{
\begin{tabular}{lccccccccc}
\toprule
\multicolumn{1}{c|}{} & \multicolumn{3}{c|}{\textbf{JPE}}                     & \multicolumn{3}{c|}{\textbf{APE}}                     & \multicolumn{3}{c}{\textbf{FDE}} \\ \midrule
\multicolumn{1}{l|}{Method} & 1.0s     & 2.0s     & \multicolumn{1}{c|}{3.0s}     & 1.0s    & 2.0s     & \multicolumn{1}{c|}{3.0s}     & 1.0s    & 2.0s    & 3.0s    \\ \midrule
\multicolumn{1}{l|}{HRI\cite{c:12}}     & 134 & 229 & \multicolumn{1}{c|}{349} & 99 & 133 & \multicolumn{1}{c|}{161}  & 93 & 177 & 295 \\
\multicolumn{1}{l|}{MSR\cite{c:29}}     & 134    & 256    & \multicolumn{1}{c|}{371}    & 97    & 142    & \multicolumn{1}{c|}{165}    &   92    & 204    & 316    \\
\multicolumn{1}{l|}{MRT*\cite{c:3}}     & 148    & 256    & \multicolumn{1}{c|}{352}    & 130   & 187    & \multicolumn{1}{c|}{218}    & 109   & 216    & 315    \\ \midrule
\multicolumn{1}{l|}{Ours*}    & \textbf{118}    & \textbf{225}    & \multicolumn{1}{c|}{\textbf{329}}    & \textbf{89}    & \textbf{132}    & \multicolumn{1}{c|}{\textbf{152}}       & \textbf{78}    & \textbf{172}    & \textbf{273}    \\ \bottomrule
\end{tabular}
}
\caption{Results of JPE, APE and FDE (in mm) on CMU-Mocap (UMPM) dataset. We compare our method with the previous SOTA methods for long-term prediction (1.0s $\sim$ 3.0s). Best results are shown in boldface. (* means multi-person motion prediction method.)}
\label{table2}
\end{table}

\section{Experiments}

\subsection{Implementation Details}
We implement our framework in PyTorch, and the experiments are performed on Nvidia GeForce RTX 3090 GPU. We train our model for 50 epochs using the ADAM optimizer with a batch size of 32, a learning rate of 0.0003, and a dropout of 0.2. For the TBPM, the kernel size and stride of 2D convolutional filter are $10\times 1$ and $stride=1$, and $padding=0$. The parameters in TRPE are: $\alpha=1,\beta=9,\gamma=\eta=2000$. The dimensions $d_z$ of keys, queries, and values in TBIFormer block and Transformer decoder are all set to 64, and the hidden dimension $d$ of feed-forward layers is 1024. There are 3 stacked TBIFormer blocks and attention layers with 8 heads in the TBIFormer and Transformer decoder. 

\subsection{Datasets}
To verify the effectiveness of TBIFormer, we run experiments on the CMU-Mocap (UMPM) dataset, which merges UMPM \cite{c:5} into CMU-Mocap \cite{c:1} for dataset expansion. Mix1 and Mix2 are blended by CMU-Mocap, UMPM, 3DPW \cite{c:2}, and MuPoTs-3D \cite{c:4} datasets. We evaluate all the methods for generalization ability by testing on the MuPoTS-3D (2 $\sim$ 3 persons), Mix1 (6 persons), and Mix2 (10 persons) datasets with the model only trained on the CMU-Mocap (UMPM) dataset. Please refer to the appendix for a thorough explanation of why we do dataset expansion and the processing detail of mixing datasets.

\subsection{Metrics of Evaluation}
\noindent\textbf{JPE Metric.} We use Joint Position Error (JPE) based on Mean Per Joint Position Error (MPJPE) to measure the poses of all the individuals, including body trajectory:
\begin{equation}\label{eq9}
\small{
{\rm JPE}(X,\hat{X})=\frac{1}{P\times J}\sum_{i=1}^{P}\sum_{j=1}^{J}{||X_j^i - \hat{X}_j^i||^2},
}
\end{equation}
where $P$ and $J$ are the numbers of people and joints. $X_j^i$ and $\hat{X}_j^i$ are the estimated and ground-truth positions of the joint $j$ for person $i$.

\noindent\textbf{APE Metric.} We remove global movement and use Aligned mean per joint Position Error (APE) to measure pure pose position error:
\begin{equation}\label{eq10}
\small{
{\rm AME}(X,\hat{X})={\rm JPE}(X-X_{r},\hat{X}-\hat{X}_{r}),
}
\end{equation}
where $X_{r}$ and $\hat{X}_{r}$ are the estimated and ground-truth root positions of human body.

\noindent\textbf{FDE Metric.} We also adopt the root position to evaluate the global movement of each person using a typical trajectory prediction metric: Final Displacement Error (FDE). The formula is described as follows:
\begin{equation}\label{eq11}
\small{
{\rm FDE}(X_{r},\hat{X}_{r}) = {||X_{r,N} - \hat{X}_{r,N}||^2},
}
\end{equation}
where $X_{r,N}$ and $\hat{X}_{r,N}$ are the estimated and ground-truth root position of final pose at $N$-th predicted timestamp.

\subsection{Baselines}
We choose 3 code-released state-of-the-art (SOTA) approaches as baselines, including two single-person based methods: HRI \cite{c:12} and MSR \cite{c:29}, and a recently released multi-person based method called MRT \cite{c:3}. HRI \cite{c:12} is an attention-based network, and MSR \cite{c:29} is a GCN-based method, which both allow absolute coordinates as input. For short-term prediction, we train all these models with 50 frames (2.0s) of input and 25 frames (1.0s) of forecasting and evaluate on the 4 datasets.  For long-term prediction, using the protocols in MRT \cite{c:3}, we set 15 frames (1.0s) of history as input to predict the future 45 frames (3.0s).


\subsection{Results }
To validate the prediction performance of TBIFormer, we follow the setting of the most single-person methods \cite{c:12,c:29} to show the quantitative and qualitative results of short- and long-term predictions, and compare our method with the baselines.

\noindent\textbf{Quantitative Results.} \Cref{table1} reports the results of JPE, APE and FDE on the 4 different datasets. Our TBIFormer significantly outperforms the baselines in prediction accuracy. We achieve up to 13\% $\sim$ 27\% improvement when compared to the previous single-person-based methods and achieve up to 13\% $\sim$ 16\% improvement compared to the multi-person-based method.~It can be noticed that MRT\cite{c:3} performs poorly in the APE metric due to the lack of spatial modeling of the human skeleton. Besides, we report the results of long-term prediction (1.0s $\sim$ 3.0s) in \cref{table2}. Our method consistently outperforms the baselines in the 3 metrics.


\begin{table}[t]
\renewcommand\arraystretch{0.9}
\centering
\resizebox{\linewidth}{!}{
\begin{tabular}{lccccccccc}
\toprule
\multicolumn{1}{c|}{} & \multicolumn{3}{c|}{\textbf{JPE}}  & \multicolumn{3}{c|}{\textbf{APE}}   & \multicolumn{3}{c}{\textbf{FDE}} \\ \midrule 
\multicolumn{1}{l|}{Method} & 0.2s     & 0.6s     & \multicolumn{1}{c|}{1.0s}     & 0.2s    & 0.6s     & \multicolumn{1}{c|}{1.0s}  & 0.2s    & 0.6s     & 1.0s     \\ \midrule
\multicolumn{1}{l|}{ w/o TBPM}     & 32    & 117    & \multicolumn{1}{c|}{195}    & 28   & 87    & \multicolumn{1}{c|}{123}    & 21   & 76    & 142    \\
\multicolumn{1}{l|}{ w/o IE, TRPE}     & 31    & 113    & \multicolumn{1}{c|}{188}    & \textbf{27}    & 85    & \multicolumn{1}{c|}{120}    & 19    & 74    & 138    \\
\multicolumn{1}{l|}{ w/o TRPE}     & 31 & 112 & \multicolumn{1}{c|}{186} & \textbf{27} & 85 & \multicolumn{1}{c|}{119} & 19 & 73 & 136 \\
\multicolumn{1}{l|}{TRPE $\rightarrow$ EuPE}     & 40 & 118 & \multicolumn{1}{c|}{191} & 34  & 89 & \multicolumn{1}{c|}{121} & 20 & 80 & 139 \\
\multicolumn{1}{l|}{ w/o SBI-MSA}    & 40    & 128    & \multicolumn{1}{c|}{208}    & 29    & 92    & \multicolumn{1}{c|}{129}    & 27    & 85    & 151    \\ \midrule
\multicolumn{1}{l|}{Full}  & \multicolumn{1}{|c}{\textbf{30}} & \textbf{109} & \textbf{182} & \multicolumn{1}{|c}{\textbf{27}} &  \textbf{84} &  \textbf{118} & \multicolumn{1}{|c}{\textbf{18}} &  \textbf{72} &  \textbf{133} \\ \bottomrule
\end{tabular}
}
\caption{Ablation studies on different components of TBIFormer. Our full method and its variants are evaluated on the CMU-Mocap (UMPM) in JPE metric.}
\label{table3}
\end{table}

\begin{figure*}[t]
\centering
\includegraphics[width=0.85\textwidth]{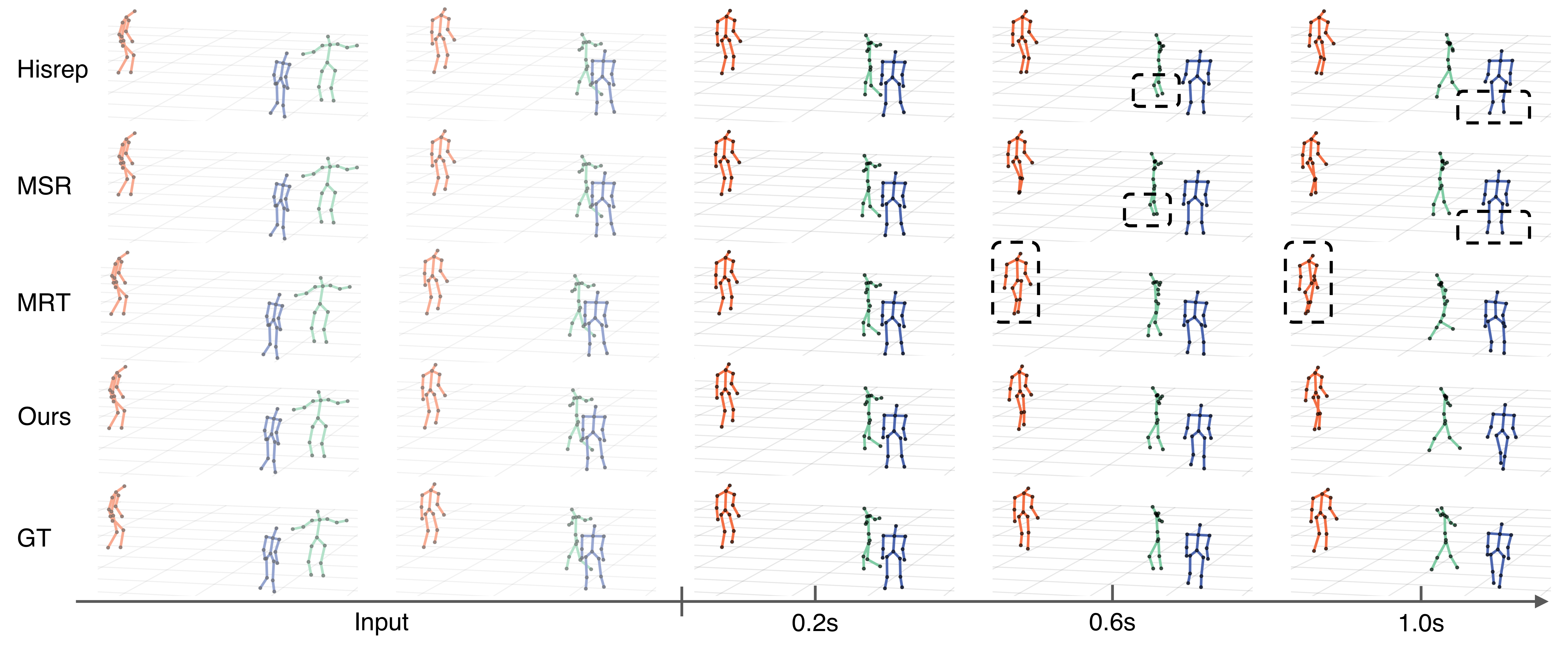} 
\caption{Qualitative comparison with the baselines and the ground truth on a sample of the CMU-Mocap (UMPM) dataset. The left two columns are inputs, and the right three columns are predictions.}
\label{fig5}
\end{figure*}

\noindent\textbf{Qualitative Results.} \Cref{fig5} shows some examples of our visualization results compared to the baselines and the ground truth. The results of HRI \cite{c:12} and MSR \cite{c:29} show that they tend to converge to a static pose in the long-term predictions. Due to the deficiency in spatial modeling of the human body,  MRT \cite{c:3} generates some distort poses. By contrast, our method generates more plausible 3D human motion in practice, which is much closer to the ground truth than others. More visualization results are supplemented in the appendix.

\subsection{Ablation Studies}
We further conduct extensive ablation studies on CMU-Mocap (UMPM) to investigate the contribution of key technical components in TBIFormer, with results in \cref{table3}. For more ablation results about the model, please refer to the appendix.


\begin{table}[t]
\renewcommand\arraystretch{0.9}
\centering
\resizebox{\linewidth}{!}{
\begin{tabular}{lcccccc}
\toprule
\multicolumn{1}{c|}{} & \multicolumn{3}{c|}{\textbf{JPE}}  & \multicolumn{3}{c}{\textbf{APE}}   \\ \midrule 
\multicolumn{1}{l|}{Method} & 0.2s     & 0.6s     & \multicolumn{1}{c|}{1.0s}     & 0.2s    & 0.6s     & \multicolumn{1}{c}{1.0s}    \\ \midrule
\multicolumn{1}{l|}{HRI\cite{c:12}}     & 51 \textcolor{red}{$\uparrow$ 2}   & 134 \textcolor{red}{$\uparrow$ 4}    & \multicolumn{1}{c|}{212 \textcolor{red}{$\uparrow$ 5}}    & 41 \textcolor{red}{$\uparrow$ 0}   & 98 \textcolor{red}{$\uparrow$ 1}    & \multicolumn{1}{c}{132 \textcolor{red}{$\uparrow$ 2}}    \\
\multicolumn{1}{l|}{MSR\cite{c:29}}     & 55 \textcolor{red}{$\uparrow$ 2}    & 149 \textcolor{red}{$\uparrow$ 3}    & \multicolumn{1}{c|}{238 \textcolor{red}{$\uparrow$ 7}}    & 46 \textcolor{red}{$\uparrow$ 0}  & 106 \textcolor{red}{$\uparrow$ 0}    & \multicolumn{1}{c}{136 \textcolor{red}{$\downarrow$ 1}}      \\
\multicolumn{1}{l|}{MRT*\cite{c:3}}    & 38 \textcolor{red}{$\uparrow$ 2} & 124 \textcolor{red}{$\uparrow$ 9} & \multicolumn{1}{c|}{203 \textcolor{red}{$\uparrow$ 9}} & 49 \textcolor{red}{$\uparrow$ 13}  & 142 \textcolor{red}{$\uparrow$ 34} & \multicolumn{1}{c}{223 \textcolor{red}{$\uparrow$ 64}}  \\ \midrule
\multicolumn{1}{l|}{Ours*}  & \multicolumn{1}{|c}{31 \textcolor{red}{$\uparrow$ 1}} & 111 \textcolor{red}{$\uparrow$ 2} & 184 \textcolor{red}{$\uparrow$ 2} & \multicolumn{1}{|c}{28 \textcolor{red}{$\uparrow$ 1}} &  85 \textcolor{red}{$\uparrow$ 1} &  120 \textcolor{red}{$\uparrow$ 2} \\ \bottomrule
\end{tabular}
}
\caption{Results on effects of random person permutation in input. All the methods are evaluated on the CMU-Mocap (UMPM) in JPE and APE metrics. The values in red indicate changes in error.}
\label{table4}
\end{table}
\noindent\textbf{Effectiveness of TBPM.} The TBPM constructs a sequence containing both temporal and spatial information for human poses. When it is removed and joint coordinates are directly concatenated for body joints in a pose sequence, TBIFormer cannot learn body part dynamics, and we can observe a significant performance decrease.

\noindent\textbf{Effectiveness of IE and TRPE.} Person identity encoding (IE) allows our method to distinguish element types in the MPBP sequence (\ie, inform each token about identity information). After eliminating IE, the model's overall performance has decreased marginally. Trajectory-aware relative position encoding (TRPE) provides ample spatial and interactive clues for the model. When we remove TRPE, the performance drops substantially. In addition, as shown from (TRPE $\rightarrow$ EuPE) in \cref{table3}, even after replacing TRPE with Euclidean distance-based position encoding, the performance is still sub-optimal. We also provide t-SNE visualization \cite{c:54} to demonstrate discriminative power between TRPE and SE (Euclidean-based encoding) in MRT\cite{c:3}. Apparently, our model equipped with TRPE can obtain more accurate and compact representations.

\noindent\textbf{Effectiveness of SBI-MSA.} The goal of the SBI-MSA is to learn body part dynamics across temporal and social dimensions. As illustrated in the final row of \cref{table3}, if the SBI-MSA is substituted with a standard self-attention module, our model only learns motion features for each person separately, resulting in poorer long-term performance. 

\noindent\textbf{Effects of Random Person Permutation.} To ensure that the people order of input data in the model should not affect its performance, we randomly permute this order during training and testing to investigate model robustness with the results in \cref{table4}. Obviously, our method is just as robust as the single person-based methods, \ie, do not rely on permutation of person in the input.

\subsection{Attention Visualization}
\begin{figure}[t]
\centering
\includegraphics[width=0.43\textwidth]{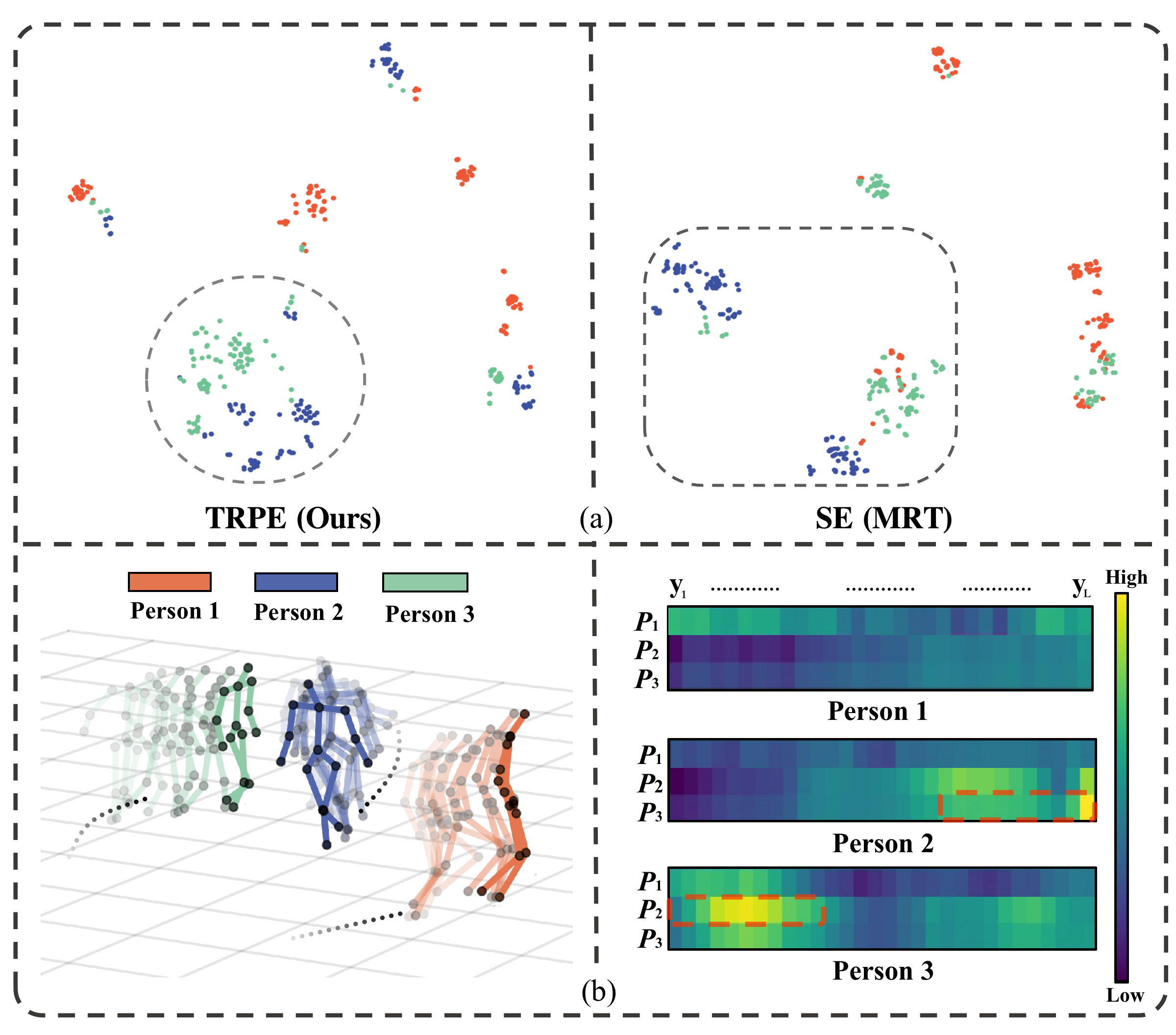} 
\caption{(a) Comparative visualization of feature distributions in t-SNE representations. The left figure shows the results obtained from our model equipped with TRPE, while the middle figure shows the results obtained from the MRT model with Spatial Encoding (SE). (b) Attention visualization of the first layer in Transform decoder. The x-axis denotes the input sequence from timestamp 1 to L, and the y-axis represents different individuals.}
\label{fig6}
\end{figure}

We show the visualization of attention score between individuals' query motion and the historical context of different people in \cref{fig6}. The left figure shows the observed motion of three people, where we can see that person 3 ($P_3$) is following person 2 ($P_2$) around, while person 1 ($P_1$) is not interacting with them nearly. The right figure draws the corresponding attention score for each individual. High attention scores for the two individuals interacting are indicated by two red-dotted regions. In terms of the high interaction group, in practice, $P_3$ should pay more attention to historical information about $P_2$ in order to adjust his behavior, which is clearly demonstrated through the visualization.

\section{Conclusion}
In this paper, we presented a novel Transformer architecture for effective multi-person pose forecasting. We first constructed a TBPM to extract spatial and temporal features based on body semantics. We also presented an SBI-MSA module to learn body part dynamics for inter- and intra-individual interactions. In addition, we proposed a novel Trajectory-Aware Relative Position Encoding for SBI-MSA to offer discriminative spatial information and additional interactive clues. Experiments demonstrated that our method outperformed state-of-the-art methods on multiple motion datasets.

\noindent\textbf{Limitations and Social Impacts.} Our work does not come without limitations. MPBP sequence involves all the individuals' body parts and time information. When inputting a long series containing many people, it will lead to heavy attentional computation during training and inference. We plan to address this issue in future. For social impacts, we are still uncertain as to whether a person can be identified based purely on his or her poses and movements. However, compared to input images of people, it is harder to invade individuals' private information.\\

{\small
\bibliographystyle{ieee_fullname}
\bibliography{egbib}
}

\end{document}